\documentclass{article}

\PassOptionsToPackage{numbers, compress}{natbib}

\usepackage[final]{neurips_2021}

\usepackage[utf8]{inputenc} 
\usepackage[T1]{fontenc}    
\usepackage{hyperref}       
\usepackage{url}            
\usepackage{booktabs}       
\usepackage{amsfonts}       
\usepackage{nicefrac}       
\usepackage{microtype}      
\usepackage{xcolor}         
\usepackage{graphicx}
\usepackage{subcaption}

\title{Exploring XAI for the Arts: Explaining Latent Space in Generative Music}  

\author{%
  Nick Bryan-Kinns, Berker Banar, Corey Ford, Courtney N. Reed, \\ 
  \textbf{Yixiao Zhang, Simon Colton,} and \textbf{Jack Armitage} \\
  School of Electronic Engineering and Computer Science\\
  Queen Mary University of London\\
  London, E1 4NS, UK \\
  \texttt{\{n.bryan-kinns, b.banar, c.j.ford, c.n.reed,} \\ 
  \texttt{yixiao.zhang, s.colton, j.d.k.armitage\}@qmul.ac.uk} \\
}

\begin{document}

\maketitle

\begin{abstract}

Explainable AI has the potential to support more interactive and fluid co-creative AI systems which can creatively collaborate with people. To do this, creative AI models need to be amenable to debugging by offering eXplainable AI (XAI) features which are inspectable, understandable, and modifiable. However, currently there is very little XAI for the arts. In this work, we demonstrate how a latent variable model for music generation can be made more explainable; specifically we extend MeasureVAE which generates measures of music. We increase the explainability of the model by: i) using latent space regularisation to force some specific dimensions of the latent space to map to meaningful musical attributes, ii) providing a user interface feedback loop to allow people to adjust dimensions of the latent space and observe the results of these changes in real-time, iii) providing a visualisation of the musical attributes in the latent space to help people understand and predict the effect of changes to latent space dimensions. We suggest that in doing so we bridge the gap between the latent space and the generated musical outcomes in a meaningful way which makes the model and its outputs more explainable and more debuggable.\\
The code repository can be found at: \url{https://github.com/bbanar2/Exploring_XAI_in_GenMus_via_LSR}

\end{abstract}

\section{Introduction}

Creating computing systems which can generate music has arguably been both a dream and a goal of researchers since the 1800s when Ada Lovelace noted that machines would one day generate ``elaborate and scientific pieces of music of any degree of complexity and extent'' \cite{Lovelace-1843}. Recently, advances in the field of generative music have relied on increasingly complex Machine Learning models  \cite{LiteratureReviewBySturm, Herremans-2017, Carnovalini-2020} -- such as neural networks \cite{todd1989connectionist, eck2002first} and deep learning techniques \cite{briot2019deep, hadjeres2017deepbach, zhu2018xiaoice, huang2018music}  -- to create convincing musical outputs. However, the complex nature of these models means that people often require some knowledge of these techniques and algorithms in order to use or adapt them effectively, making them difficult for people, especially non-experts, to understand and debug. The presentation of these models in current interactive musical systems also means that much of the generation process is invisible to the user; very few musical applications provide an interface which allows the user to visualise how a piece of music has been created or explain how their interaction affected the musical content. This means many generative systems, and digital musical instruments in general, leave artists feeling disconnected from their work, or worse, are generally inaccessible to musicians or anyone besides the creator \cite{wallis:engagement, benyon:designSE, morreale:longevity}. 

In this paper, we explore how eXplainable Artificial Intelligence (XAI) can be applied to generative music systems to both aid with human understanding of the model and to support inspection and debugging of the model and its outputs. First, we outline an overview of XAI and current applications within the arts, followed by a summary of a systematic literature review of the explainability of 87 creative AI papers. Then we introduce our implementation of XAI for the arts by demonstrating how the latent space of a generative music model can be made more explainable -- we contribute a novel user interface which supports real-time navigation of the latent space of a generative music model and includes the generation of a set of piano rolls, colour plots, and audio files, for a trained MeasureVAE \cite{pati2019learning} model.
This is achieved by sampling the latent space from MeasureVAE and regularising the dimensions for a set of musical metrics \cite{pati2, pati2020attributebased}.
We conclude by reflecting on the implementation, and we suggest future directions for research on XAI for music and the arts.



\section{Related Work}

The field of eXplainable AI examines how machine learning models can be made more understandable to people, thus increasing their usability and making it possible for non-experts to utilise them in a variety of contexts. In particular, XAI researchers explore how non-intuitive and difficult-to-understand AI models  -- such as neural networks and deep learning techniques -- can be explained \cite{DARPA, gunning_2016}. For example, XAI projects have focused on creating human-understandable explanations of why an AI system made a particular medical diagnosis \cite{quellec2021explain}, how the AI models in an autonomous vehicle work \cite{du2019look, shen2020explain}, and what data an AI system uses to generate insights about consumer behaviour \cite{sajja2021explainable}. These XAI applications improve human-in-the loop use, making it easier to integrate AI into every day tasks and improve the accuracy of systems in combination with the expertise and human intelligence of the user.

Having more explainable AI for the arts is important for AI arts systems that we co-create with, referred to as \emph{creative AI} models, as both artists and audiences would benefit from a  better understanding of i) what an AI model is doing to generate artistic content and ii) why this artistic content was generated in response to their own artistic input. This frames the AI as a tool for creating and co-creating content, rather than a mysterious and opaque box of tricks, uncontrollable by the artist. Indeed, having more transparent and understandable AI models is essential for creative AI as co-creation implicitly requires some level of mutual understanding and engagement both with the artistic output and with each other \cite{louie2020cococo}. Co-creating with an AI requires us to be able to inspect, understand, modify and debug both the AI model and the output it creates in response to an artist's input. In this way, the artist can understand their impact on the system and experience feelings of ownership over their artwork. Designing for this level of understanding of what an AI system is doing is a key focus of the field of XAI. 
More broadly, this relates to the notion of \emph{framing} information presented in \cite{charnley2012notion} and surveyed in \cite{cookframing}, whereby a generative AI system describes its creations with additional text or other stimuli. This has been taken further in \cite{llanoxcc}, with the suggestion that a creative AI system should engage in dialogue with users to convince them of the value of its output. 
 
The arts, and especially music, also provide a complex domain in which to test and research new AI models and approaches to explainability. Compared to domains such as healthcare and automotive industries, the arts require similar levels of robustness and reliability from their AI models, but have significantly fewer ethical and life-critical implications, making the arts a great test-bed for AI innovation. In other words, exploring approaches to XAI for the arts could both inform the design of XAI for more safety-critical systems and lead to more intuitive and engaging co-creative AI systems. For example, music interaction provides an opportunity to study a system's sensitivity to time-critical parameters since real-time, understandable feedback is critical for musicians in co-creating with digital instruments \cite{jack:latency, mcpherson:latency}.

As current XAI research is predominantly focused on functional and task-oriented domains, such as financial modelling, it is difficult to directly apply XAI techniques to creative AI models.
Moreover, the majority of papers about XAI beyond simple explanations do not focus on the implementation of XAI, but instead offer design guidelines for explainability \cite{Liao-2020} or theories of how such explainability might work \cite{andres2020scenario}. This means that there are few explainable AI systems to build from. To compound this problem, there are few creative AI models which provide any explanation of how the model works or expose any elements of the creative AI model in any meaningful way, which we demonstrate for the domain of generative music in the following section.  
 
\subsection{XAI and Generative Music}\label{sec_xai_and_generative_music}

Taking music generation as a key example of creative AI models, we surveyed 87 recent AI music papers from venues including the New Instruments for Musical Expression Conference (NIME) series and the Computer Music Journal to examine what role the AI had in the creative process and how much of the AI in these interactive systems is actually explained or exposed to humans in the system itself (rather than being explained in the paper). Our review sample started with the 94 papers reviewed in \cite{Herremans-2017}; 19 papers which were purely related to cognitive theory or not accessible to the authors were removed. We also added 12 more recent creative AI papers to represent current practice, including the use of XAI in other HCI fields \cite{sturm:AIlit, hazzard:AIlit, andres2020scenario, long:AIlit, malsattar:AIlit, Benetatos:AIlit, Proctor:AIlit, Lupker:AIlit, Zhang:AIlit, roberts:AIlit, Gillick:AIlit, Louie:AIlit}, making a total of 87 papers reviewed. We analysed the papers using three existing frameworks to capture the key features of \emph{XAI for the arts}. Specifically, we examined the following for AI in the interactive arts: the \emph{role} of the AI in co-creation, possible \emph{interaction} with the AI, and how much \emph{common ground} humans might be able to establish with the AI, as follows.

\emph{The role of the AI} – we used Lubart’s classification of the role of AI as a creative partner \cite{lubart-2005}
to classify the role of the AI from models which take care of generative tasks without interacting with humans through to AI models which take on the role of a colleague in creative collaborations.

\emph{Interaction with the AI} – we used Cornock and Edmonds’ classification of interaction styles with interactive art from static to dynamic-interactive \cite{Cornock-1973} as the more interactive and responsive an AI is, the more chance there is for people to understand what the AI is doing and might do in the future. 

\emph{The common ground with the AI} – we drew on Clark and Brennan’s work on grounding in human communication \cite{Clark1992} to classify what a person might be able infer about an AI’s output state from a low stage of grounding where a person is only aware that some output has been made through to a high stage of grounding where they have an understanding of its meaning and can make an informed reaction to the output.

Our perspective is that the explainability of creative AI is a combination of the role the AI takes, the interaction it offers, and the grounding that can be established with the AI. The more collaborative the role is, the more explanation is required which in turn necessitates more interaction and grounding. Increased opportunities for interaction help people to learn about and infer an understanding of the AI and its behaviour. Increased levels of grounding offer more chances for people to understand what a creative AI did, and why. Increased interaction and grounding offer more changes for people to inspect, understand, and debug AI models and their creative output.

In our review, we found some excellent examples of creative AI which take the role of a colleague. For example, Shimon the robotic marimba player \cite{hoffman2010shimon} listens to live human players, analyses perceptual aspects of their playing in real-time and plays along in a collaborative and improvisatory manner. In terms of interaction, Shimon provides a real-time feedback loop within the art work itself, making the collaboration highly dynamic and interactive. However, Shimon offers only mid stage grounding as there is no explanation of what it did, nor \emph{how} or \emph{why} Shimon made particular musical responses in the improvisation. 

Other generative tools, such as Hyperscore \cite{farbood:hyperscore}, demonstrate higher levels of common ground with the user. In Hyperscore's interface, the reactive change to input is displayed in a piano-roll notation. This allows the user to observe the effects of their input and develop an understanding of the system's response. The interactive controls in the interface provide a way for the user to experiment with melody creation, while the system preserves their original melodic curve ideas and allows them to make incremental and reversible changes. In this way, the user can see \emph{how} their input changes the output; however, the internal model and the reasons \emph{why} the system reacts are still largely obscured.

In our review, we found that 73 of the 87 papers took the role of generating music without any human collaboration. There was also little interaction offered by the AI music systems we reviewed: 41 papers did not not offer any real-time interaction with humans at all, but rather generated melodies from training data, without any user input or decisions. Additionally, 76 papers were at the lowest stage of grounding meaning that although the AI makes a musical contribution, a person would not reliably be able to discern what the AI system did based on their input, and would simply be aware that some musical output was generated somehow. These kinds of creative AI are difficult to debug as their implementations are not meaningfully exposed, and the technical complexity of their interfaces prevents musicians from using them.

To summarise, while there are compelling examples of creative AI systems that collaborate with musicians, only a few creative AI systems explain what their models are doing, how they do it, and why. This makes debugging such models and their output problematic. The rest of this paper explores this gap through our approach to explainable latent spaces in music generation.


\section{Explaining Latent Spaces}\label{explaining_latent_spaces}
As our first step in investigating more explainable AI for the arts we have been researching how to make latent spaces in AI music models more understandable.
Recently, some systems developed for human-AI co-creation have exposed the latent space of generative music models \cite{louie2020cococo, pati2019learning, Thelle2021Spire, Murray2021Latent, gold29044}, meaning that users can create their own music by navigating a possibility space, typically represented on a blank 2D grid. For example, latent space models have been successfully used for creative applications such as music generation \cite{yang2019deep, wang2020pianotree}, music inpainting \cite{pati2019learning}, and music interpolation \cite{roberts2018hierarchical}. However, to date, these systems have been opaque AI models and do not offer the higher-level XAI properties of interaction and grounding we discussed in Section \ref{sec_xai_and_generative_music}.
For example, Murray-Browne and Tigas \cite{Murray2021Latent} trained a latent space on a set of dance postures which was then mapped to various musical outputs; although dancers found distinctive ways of performing with the system, how their movements directly influenced the music was unclear.
If we can offer more explainable approaches to exposing latent spaces then there is more chance for people to be able to debug the AI model, its training, and the output it generates.

For our system, we built on the popular MeasureVAE system\footnote{\cite{pati2019learning} and \cite{pati2}, which we build upon in our work, are licensed under a Creative Commons Attribution-NonCommercial-ShareAlike 4.0 International License.},  which Pati \textit{et al.} \cite{pati2019learning} describe as being ``successful in modeling individual measures of music''. In response to an input extract of music, the model generates a similar measure of music by: i) encoding the input measure into a latent space via bi-directional recurrent neural network (RNN), ii) sampling the encoder's latent space (\textit{z}), and iii) decoding \textit{z} via a combination of RNNs and linear stacks. Notably, the decoder uses two uni-directional RNNs: one is responsible for beats (four beats in a measure), and the other one is for ticks (six ticks/ symbols per beat) \cite{pati2019learning}. For a full description, we point the reader towards \cite{pati2019learning}.

\begin{figure}[h]
  \centering
  \includegraphics[scale=0.25]{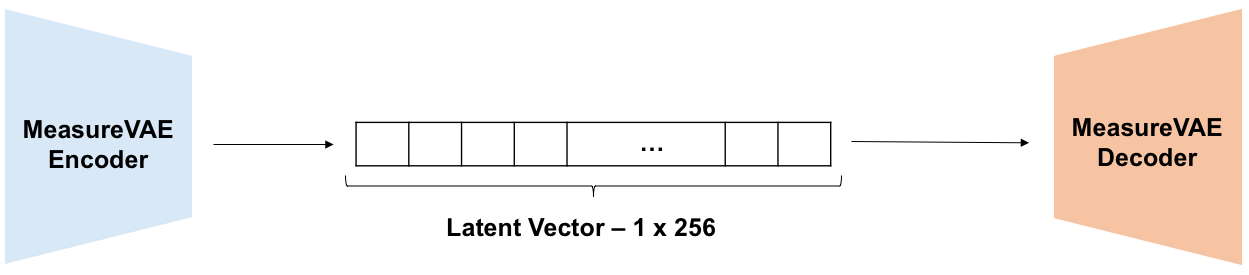}
  \caption{The simplified MeasureVAE}
 \label{fig:VAEwithoutLSR}
\end{figure}

In keeping with current research on AI music generation we trained MeasureVAE using publicly available music from a single genre to ensure coherence of the training data - in our case 20,000 publicly available monophonic Irish folk melodies \cite{Sturm2016MusicTM}. The data is partly anonymous, showing contributor's names from a community website dedicated to Irish traditional music, and contains no offensive content. This produces a latent vector of 256 dimensions, as illustrated in Figure \ref{fig:VAEwithoutLSR}. Once trained, inputting a melody into the encoder will generate a new melody through the decoder. The features of the new melodies can be varied by modifying the values of the dimensions in the latent vector. However, there is no way for a person to know what effect changing these dimensions would have on the generated music. In addition, the 256 entangled dimensions make it difficult to perceive a difference in output from a change in any single dimension, making our Irish folk song generator opaque and unexplained.

\begin{figure}[h]
  \centering
  \includegraphics[scale=0.25]{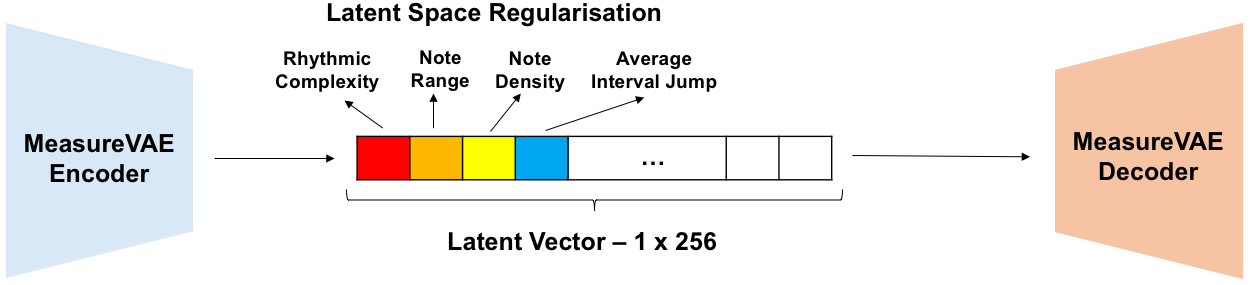}
  \caption{The simplified MeasureVAE with LSR}
  \label{fig:VAEwithLSR}
\end{figure}

However, if we use latent space regularisation (LSR) \cite{hadjeres2017glsrvae} in training the VAE -- which has been widely used in the study of controlled generation of images \cite{lample2018fader} and music \cite{pati2, tan2020music} -- we can make this creative AI approach more understandable and explainable. We use LSR to force some specific dimensions of the latent space to represent specific musical attributes similar to the method in \citet{pati2, pati2020attributebased}. Specifically, we assign dimensions \textit{0}, \textit{1}, \textit{2} and \textit{3} to rhythmic complexity, note range, note density, and average interval jump respectively (see Figure \ref{fig:VAEwithLSR}). We selected these metrics as typical examples of meaningful musical features used in current research in order to demonstrate our approach.

In this way, specific dimensions in our latent space correspond to meaningful musical attributes in AI-generated outputs. Since these attributes are manipulable, they can form the basis for a real-time user interface illustrated later in this paper, thereby creating a feedback loop between input music, human modifiable dimensions, and AI-generated music. We suggest that this feedback loop can support the debugging of the creative AI and its output. Furthermore, we suggest that this support can be offered in a form that is commensurate with musician's existing skills and usability expectations.

\subsection{Implementation} \label{sec_implementation}

 To utilise the MeasureVAE architecture with the latent space regularisation technique, we build on the implementation of \cite{pati2}\footnote{\url{https://github.com/ashispati/AttributeModelling}}. The training data of 20,000 monophonic Irish folk melodies \cite{Sturm2016MusicTM} are converted into a measure-based data representation. Each measure is represented with 24 characters, where each character corresponds to one of the sixteenth note triplets in a 4/4 measure, with a total of 24 characters per measure \cite{pati2019learning, pati2020attributebased}. These characters include note names (A3, G5, ...), continuation tokens (\_) and special rest tokens.

We use four musical attributes in our model following \cite{pati2,pati2020attributebased}; Toussaint's rhythmic complexity \cite{Toussaint02amathematical}, note or chromatic range (\textit{max pitch - min pitch}), note density (number of notes in a measure) and average interval jump. Average interval jump represents the average of the absolute values of the interval between adjacent notes in a single measure melody.

We jointly train our MeasureVAE model with latent space regularisation on all four attributes, to force four specific dimensions of the latent space to represent given musical attributes \cite{pati2, pati2020attributebased}. We apply these constraints to the first four dimensions of the 256-dimensional latent vector, and assign them to rhythmic complexity, note range, note density and average interval jump, and add an attribute-specific regularisation loss to the training objective of the VAE. Specifically, for each attribute, a musical metric value (e.g. average interval jump) is calculated for each item in a mini-batch. Then, a distance matrix (\textit{D$_{attribute}$}) is created by calculating the distance between each item's metric value and all the other item's metric values resulting in an \textit{N x N} matrix (\textit{N} examples in a mini-batch). Similarly, another distance matrix (\textit{D$_{dimension}$}) is created for the values of the regularised latent dimension, again resulting in an \textit{N x N} matrix. Subsequently, the mean squared error of \textit{(tanh(D$_{dimension}$) - sgn(D$_{attribute}$))} is added to the VAE objective. Finally, after training the model with LSR, the values of these dimensions become monotonically tied to the corresponding music attributes. Therefore, when we change the values, the corresponding attributes of the generated music change accordingly.

We use Adam \cite{kingma2014adam} as the optimizer of the model with learning rate = 1e-4, $\beta_1$ = 0.9, $\beta_1$ = 0.999 and $\epsilon$ = 1e-8. The model is trained on a single GeForce RTX 2080 Ti GPU for a total of 30 epochs following the same setting of \cite{pati2}, taking an average of 2.5 hours per epoch. After training, the reconstruction accuracy of the LSR model achieves 99.87\% on the training set and 99.68\% accuracy on the validation set, and the non-LSR model achieves 99.84\% and 99.77\% respectively\footnote{The non-LSR model is trained for 11 epochs as this gave the best accuracy.}. As calculated in \cite{pati2, pati2020attributebased}, we have an interpretability metric, which is from \cite{adel_interpretability} and measures how well we can predict an attribute using only one dimension in the latent space. For the LSR model, interpretability scores for rhythmic complexity, note range, note density and average interval jump are 0.80, 0.99, 0.99 and 0.91 (average 0.92) for their corresponding dimensions, respectively (the higher the better). For the non-LSR model, interpretability scores are 1.5e-4, 9.1e-6, 1.7e-5 and 1.2e-6 in the same order.

Two user interfaces (UIs) were built using React.js and deployed as web applications online to demonstrate interaction with LSR\footnote{\url{https://xai-lsr-ui.vercel.app/}} and without LSR\footnote{\url{https://xai-no-lsr-ui.vercel.app/}}.
In each demonstration, we encode an input MIDI measure using the trained encoder and obtain its latent vector. Then, to generate musically controlled variations of it, we manipulate the values of the first four dimensions and decode these manipulated latent vectors to obtain output music sequences. To demonstrate the explainable latent space, we sweep the regularised latent dimensions discretely and sample values. For each dimension, we take 10 equally spaced samples, creating every possible latent vector combination using these 10 samples per regularised dimension, in total 10,000 latent vectors for four dimensions / attributes. In these latent vectors, values for non-regularised dimensions are kept as they are in the latent vector of the encoded input sequence. To determine the sampling limits, we store the latent vectors of the training set and generate histograms of the regularised dimension values. The limits are then set according to the histograms (with slight modifications to allow wider ranges in terms of the musical metrics). Once we have these 10,000 latent vectors, they are decoded into output music sequences and we generate MIDI files of them. For the sake of practicality, we generate piano-rolls and MP3 files for each MIDI output in advance, and host them online for the web demos. The generated MIDI, piano-roll and audio files are available at our GitHub repository\footnote{\url{https://github.com/bbanar2/Exploring_XAI_in_GenMus_via_LSR}}. 

\subsection{Interaction}\label{sec_interaction}

\begin{figure}
  \centering
  \includegraphics[scale=0.22]{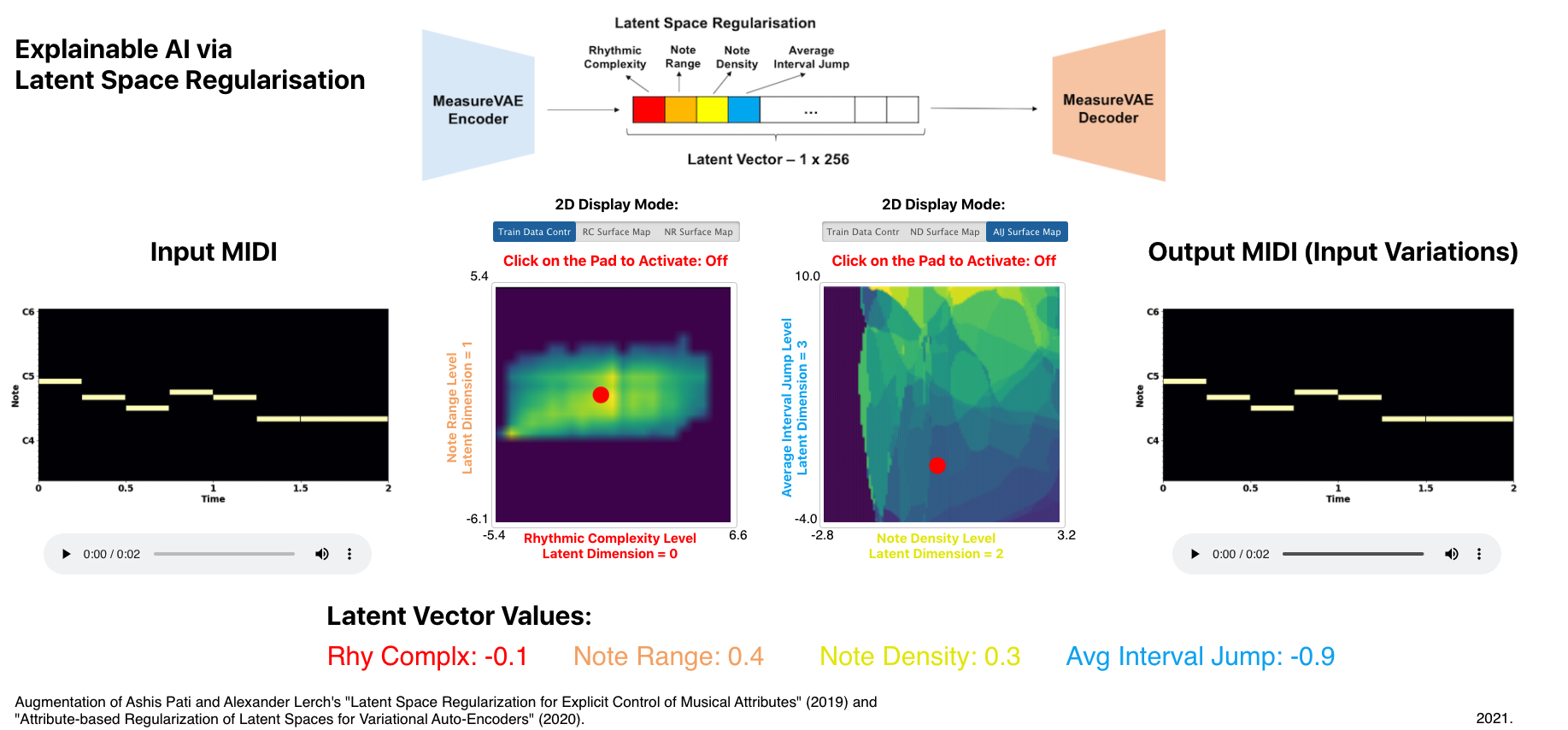}
  \caption{Screenshot of the user interface with LSR}\label{fig_LSR_UI}
\end{figure}

Both demos feature UIs that provide real-time interaction with our creative AI system. For example, Figure \ref{fig_LSR_UI} shows the UI for the demo with LSR. In both UIs the input MIDI measure is shown and can be listened to on the left-hand side of the UI in piano-roll format, and the generated output MIDI measure is shown and can be listened to on the right-hand side of the UI also in a piano-roll format. The main points of interaction are two 2D-pads in the centre of the UI; when clicked on, the user can navigate the pads by dragging their mouse, which controls the red dot. The left-hand pad controls the rhythmic complexity and note range dimensions, whereas the right-hand pad controls the note density and average interval jump dimensions (see the axis labels on the LSR version in Figure \ref{fig_LSR_UI}). These two pairings were chosen to maximise the semantic differences between the dimensions in each pad in order to offer users more observable effects of changes in position within the pad. 

The white dots on the 2D pads refer to the corresponding latent vector dimension values of the input sequence, given as reference points. As users hover over these pads -- selecting different latent dimension values -- the output MIDI is updated in real-time and played back to people. Musically, these outputs correspond to variations of the input sequence which vary as we manipulate its latent vector for generation; for example, we will have a higher range of notes in the generated musical sequence when the note range dimension is high.

In our demo with LSR we increase the explainability of the AI in two ways: i) key parts of the AI model are exposed to the user in the interface and meaningfully labelled (in this case, with musical features), and ii) the real-time interaction and feedback in the UI allows people to explore the effects of these features on the generative music and thereby implicitly learn how the model works.

Referring to the three properties of XAI for the arts described in Section \ref{sec_xai_and_generative_music}, in our implementation the AI acts somewhat like a colleague \cite{lubart-2005} -- the response to the user is given in real-time, as would be done in a human-to-human musical interaction. This drives a feedback loop between a user and the AI, whereby a person's reaction to the AI's response informs the subsequent interaction. Thinking musically, this resembles a duet in creative improvisation, where the players make real-time decisions based on their colleague's performance. 

In terms of grounding between the AI and the user, the system displays the possible user input parameters to the model through the labeled pads, 2D displays, and movable cursors, and moreover, the visualisation changes in response to these control actions. As the UI provides musically meaningful labels and an interaction feedback loop (which allows for exploratory user learning to predict how user input might change the AI's output), we see this system as being towards the highest level of grounding \cite{Clark1992}, where a person is provided with cues to the AI model's current state and can predict possible next steps. Furthermore, the real-time interaction creates a sensation of ``playing'' the model and helps to recreate other familiar musical interfaces through the use of the piano-roll notation. Real-time feedback provides musicians with an assurance that their input is being received, increases accuracy in timing during use, and positively influences their perceptions of the quality and usability of a system \cite{jack:latency}. Comparing this to other generative music systems which often take input at the command line, the use of pads and sliders and note visualisation on a piano-roll (commonplace in digital musical interfaces) is more intuitive and typical of musical interaction. In this sense, the system provides an interaction which is both dynamic and understandable in terms of the generative system itself and also tailored around the specific context of music creation. 

\subsection{Visualisation}

\begin{figure}[h]
\label{visualisations}
\begin{subfigure}{.16\textwidth}
  \centering
  \includegraphics[width=.95\linewidth]{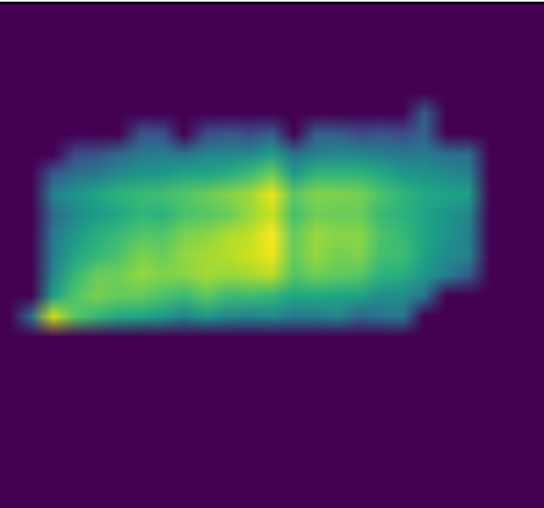}
  \caption{TDRCNR}
\end{subfigure}
\begin{subfigure}{.16\textwidth}
  \centering
  \includegraphics[width=.95\linewidth]{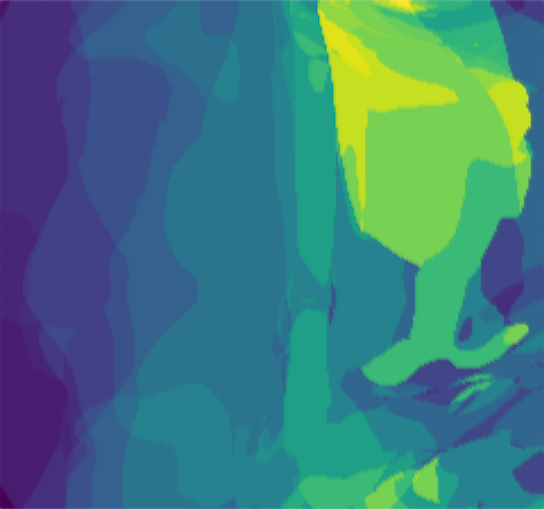}
  \caption{RC}
\end{subfigure}
\begin{subfigure}{.16\textwidth}
  \centering
  \includegraphics[width=.95\linewidth]{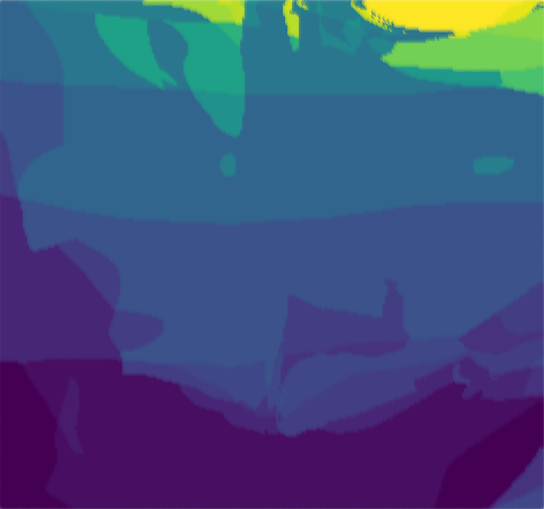}
  \caption{NR}
\end{subfigure}
\begin{subfigure}{.16\textwidth}
  \centering
  \includegraphics[width=.95\linewidth]{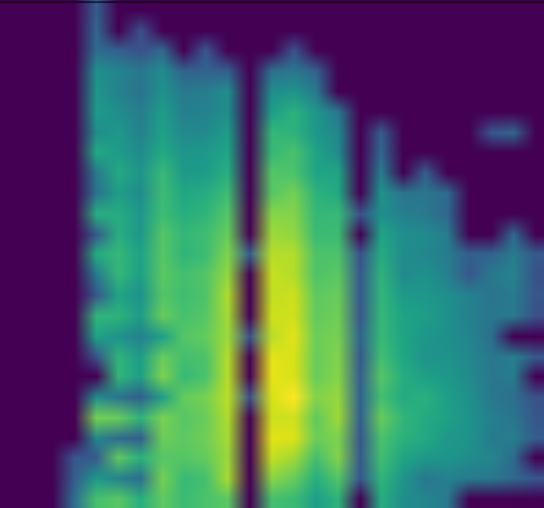}
  \caption{TDNDAIJ}
\end{subfigure}
\begin{subfigure}{.16\textwidth}
  \centering
  \includegraphics[width=.95\linewidth]{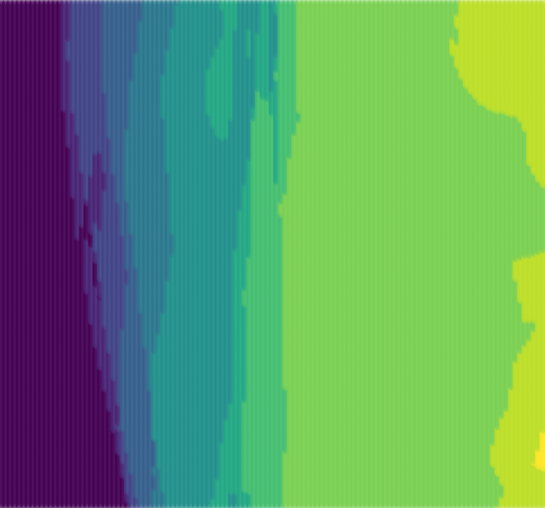}
  \caption{ND}
\end{subfigure}
\begin{subfigure}{.16\textwidth}
  \centering
  \includegraphics[width=.95\linewidth]{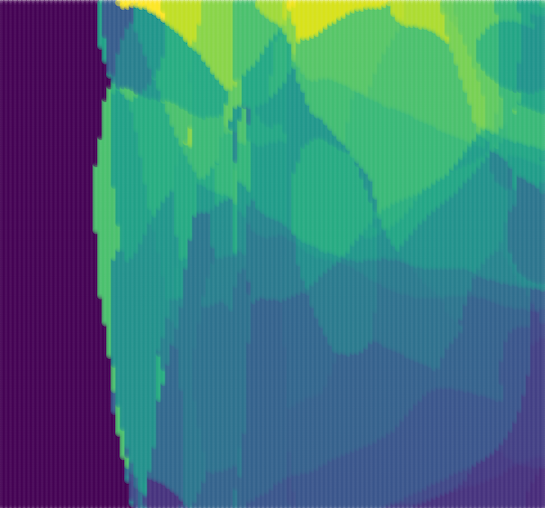}
  \caption{AIJ}
\end{subfigure}
\caption{Visualisations: a) Training Data visualised in terms of rhythmic complexity and note range (TDRCNR); b) Rhythmic Complexity surface map (RC); c) Note Range surface map (NR); d) Training Data visualised in terms of note density and average interval jump (TDNDAIJ); e) Note Density surface map (ND); f) Average Interval Jump surface map (AIJ)}\label{fig_LS_vis}
\end{figure}

To provide further insight into our AI model, we display three 2D plots within each of the pads based on the visualisations in \cite{pati2} as illustrated in Figure \ref{fig_LS_vis}.
Firstly, training data contribution plots (\ref{fig_LS_vis}a and \ref{fig_LS_vis}d) for the left and right-hand pads respectively represent how many items in the training data set have contributed to a specific location in the latent space. Statistically, this provides an idea of how unlikely the output will be for any location, given the musical structure of the dataset. In each of the pads we also have two surface map figures as used in \cite{pati2, pati2020attributebased} and illustrated in \ref{fig_LS_vis}b and \ref{fig_LS_vis}c for the left-hand pad, and \ref{fig_LS_vis}e and \ref{fig_LS_vis}f for the right-hand pad. We create these surface maps by decoding corresponding latent vectors for each point (non-LSR dimensions are kept the same as the encoded input vector, and the other two LSR dimensions are taken from the latent vector of the input music) and calculating the musical attribute values for each decoded sequence. Since we have two attributes per 2D pad, we have two surface maps for each pad. These surface maps illustrate how the LSR technique works since the metric values increase (yellow regions) for the higher parts of the corresponding axis. These surface maps also offer detailed information about regions to hover over, and allow inference of the potential kinds of outputs resulting from different points in the latent space dimensions. In doing so we offer a different kind of explainable interaction than other creative AI systems which allow interaction with latent space such as \cite{gold29044, Thelle2021Spire, Murray2021Latent}. In other words, we increase the explainability of the UI by providing a rich visualisation from which people can infer, with some meaning, the gist of what the AI model might generate for particular latent dimension values as they interact with it.

The non-LSR demo UI surface maps are different to the LSR version and show the same musical attribute value for any point in the latent space. This is because there is nothing meaningful about these dimensions as LSR has not been applied and changing the value of these individual dimensions in a 1 x 256 vector does not have any significant effect on the decoded sequence (which is to be expected behaviour for this non-LSR case). In other words, simply visualising the surface maps in the non-LSR version does not increase the explainability of the AI. 

\subsection{Debugging}

By using latent space regularisation to force a small set of dimensions to be mapped to semantically meaningful musical features, we hope to better support artists using MeasureVAE in key debugging activities of \emph{inspecting} the model, \emph{understanding} how the model works, and \emph{changing} the model to create the desired output.

We suggest that the real-time nature of the feedback provided by our demo would make it easier to inspect and develop an understanding of the AI model. The user is immediately able to see the results of their actions, which informs any follow-up from the user needed to debug the system or their previous input. 
While this is an important feature for debugging in general, it is especially the case in a music context: musical interaction is time-dependent, meaning an artist must quickly be able to make an informed reaction or otherwise risk derailing a performance or composition process.
By supporting real-time feedback, we increase the common ground between the AI and the human. This elevates the creative process from a situation where we know that the AI has done something in response to our input, but we are not sure what (e.g. Shimon \cite{hoffman2010shimon}), to a level where we can directly see and start to understand what the AI has done in response to our input. We suggest that users would thus be able to predict (with some learning on the human side) how changes to the regularised dimensions of the latent space change the generative output in a way which can better fulfil their musical intentions. 
It is worth noting that there are many other time-sensitive applications where such XAI features would greatly benefit the user, including the more functional and task-oriented tools described earlier in medical care and transportation.

Importantly, our demo retains its original input whilst other parameters are changed. This allows users to compare the current AI output with their original contribution, contributing to debugging of expected outcomes. This is similar to the design of Hyperscore \cite{farbood:hyperscore}, where a melodic curve is always retained as reference, whilst other parameters are changed. Moreover, in our demonstration, users are able easily revert back to previous settings by moving the red dots in the UI (Figure \ref{fig_LSR_UI}) between different positions in the latent space. Users are also able to quickly observe the results of their tinkering in real-time, and so may develop a better understanding of the mapping between changes in latent space dimension values and the resultant generated output. This form of interaction allows for trial-and-error debugging of the creative output as well as supporting users in exploratory learning of dimension mappings. In this way, we support creativity support tool design principles such as Shneiderman \textit{et al.}’s \cite{Shneiderman2006} principles of supporting exploration (as people can quickly trial different ideas), and offering a `high ceiling’ of tools (as people can tinker with a wide range of different options).

In summary, combining support for inspecting and understanding of the AI model with being able to interactively manipulate the model offers us opportunities to debug the model. In other words, by providing more grounded and interactive explanations of what the AI is doing we offer greater opportunities for creative exploration of the musical space, and importantly, increased support for debugging of both the creative output and the AI model itself.



\section{Limitations, Societal Impact, and Future Work}\label{sec_implications}

There are currently several limitations to our demo which need to be addressed in further work. Firstly, the musical parameters chosen for interaction with the system are a somewhat arbitrary selection of parameters typically found in contemporary research on AI music generation and present only a small subset of the variables in music creation. Future work needs to explore which features are most useful for explaining the workings of the AI model.

Secondly, the system and interaction focus on manipulating the pitch and rhythmic aspects of a single measure of music. This ignores timbre and higher-level structure in music, operating with only a fraction of the variables musicians have control over on a traditional instrument or composition tool.

Thirdly, for a non-musical user, the system may still present some explainability barriers. The chosen parameters -- although labelled and visualised -- require the user to have some prior understanding of music. Other control mechanics, such as the semantic sliders used in \cite{louie2020cococo}, may be more appropriate, although they currently do not give explanations for the link between interaction and output.

Fourth, our surface map visualisations are static and intended to be illustrative of the regularised dimensions. Future work could investigate dynamic surface maps which are rendered according to the currently selected points in the dimensions e.g. the surface map for the note density \& average interval jump 2D pad could be changed in real-time according to the currently selected point in the rhythm complexity \& note range 2D pad.

Fifth, we only focus on explainability of the latent space forming at the bottleneck of the MeasureVAE architecture, in-between the encoder and decoder blocks. To extend explainability to the layers of encoder and decoder blocks, one technique we are interested in applying is Concept Whitening (CW) \cite{Chen_2020}. CW can be practically applied to any layer of a network and could be used to demystify how the network learns concepts through those layers without harming the main training objective. This technique aligns concepts, which in our case might be musical attributes or more high-level concepts such as genres or moods, with latent space axes of a given layer by doing de-correlation, standardisation and orthogonal transformation \cite{Chen_2020}.

Sixth, the comparison of the LSR version to the non-LSR version of the demo might be improved by using some non-semantic dimensionality reduction technique in the non-LSR demo to address the lack of perceptible effect of user changes to dimensions in the non-LSR demo. For example, the non-LSR demo could be extended to compute Singular Value Decomposition (SVD) in the latent space over the training set and then allow the user to manipulate the four SVD directions with the highest variance. This would allow comparison of two demos in which user interaction has perceptible effect on both demo's musical output, allowing us to compare the explainability of semantically regularised dimensions versus non-semantic dimensionality reduction.

Finally in terms of implementation, the novelty of our implementation demo is limited as it involves the combination of two published approaches. Future work needs to explore a larger set of combinations of approaches drawing from both published and unpublished research.

Our current system suffers from a problem commonly found with AI music models: the dataset used to train the model is limited within the Western music canon, and even more so to a specific genre of folk songs. This limits the output to specific tonal and rhythmic features. It is possible that the compositions created using a tool like ours will sound similar to the songs of the dataset we use, which decreases the diversity of music created. Future research should explore the utility of our approach with training sets of different genres of music, and also training sets containing combinations of genres.

The system presented in this paper needs in-depth study and validation by users with a range of musical knowledge. For example, we selected two pairings of dimensions for the 2D pads in the centre of the user interface as discussed in Section \ref{sec_interaction}. The validity of this design choice needs to be evaluated with users in terms of: i) whether the chosen pairings (left-hand: rhythmic complexity \& note range; right-hand: note density \& average interval jump) are the most useful and intuitive for users; and ii) whether including additional 2D pads would improve the explainability of the system or instead be confusing for users e.g. six 2D pads could be displayed to users to present all possible combinations of the four LSR dimensions but this would substantially increase the visual complexity of the user interface and may lead to cognitive overload.
In addition, further development of such a system must include working directly with potential users on how such a system may benefit their composition or performance practice, and how their own artistic identity can be better incorporated and expressed. Moreover, explorations of use cases by artists will further suggest interaction mechanics and changes to the user interface, and provide insight into how the system might assist in creative and debugging processes.

As we outlined earlier in this paper, undertaking AI research in the arts provides a demanding real-world and real-time context in which approaches such as XAI can be explored, without the risk of substantial negative societal impact in life-critical domains such as healthcare and automotive industries.
However, there are potential negative societal impacts on artistic practice and livelihoods of our work and creative AI research in general which must be considered.
Most importantly, some argue that co-creative AI would diminish human creativity, remove the human from the creative process, and devalue human creativity itself. There is also concern that the use of generative models would lead to a homogenisation of music and a marginalisation of musical skills and traditions which are not amendable to reproduction by AI. In our view, through human-centred XAI and the design of UIs which embrace the user's role and interaction with the AI model, as presented here, we can proactively work to ensure that the artist remains key to the the creative process. Indeed, our view is that working with artists to design and implement explainable AI systems will help to mitigate concerns about the impact of AI on creativity.

\section{Conclusions}
Explainable AI is a growing research field which has the potential to contribute to making AI systems more co-creative. Taking AI music as a key example of AI for the arts, it is clear that there is huge potential for more explainable AI models, given the limited explainability of current creative AI models. Typically, these offer limited interaction and low levels of grounding between AI and human -- a situation where we notice that the AI has created something, but are not sure how the AI output relates to our input. In this paper, we demonstrated how latent spaces can be made more explainable, and in doing so showed how they could support debugging as part of creative practice.

We suggest that future work in XAI for the arts should move away from the functional explanations explored by current XAI research and focus instead on conveying the \emph{gist} of what AI models are doing. Much like the third wave of Human-Computer Interaction (HCI) \cite{Bodker2015} which shifted HCI research to focus on experience and meaning making, conveying gist will be a paradigm shift in how we design and use AI in creative settings. By following an interdisciplinary approach, where creative AI presents information that is meaningful to people -- such as by presenting visual cues between mappings \cite{gold29044}, or visualising levels of mutual trust with emoticons \cite{McCormack2019} -- we can better support human-AI collaboration. Once we create AI systems that convey the gist of what they are doing creatively, we will have the chance to mutually engage with AI in co-creation.

Finally, we believe that when we think of ``explainable'' AI, we should consider how the design of our systems embrace the existing knowledge, experience, and practices that users will bring to the interaction cf. \cite{Liao-2020}. A critical question designers can ask is whether the AI is explained in the context in which it will be used. In this paper, we present an interactive music generation system which works with interface elements familiar to musicians, and focuses around the real-time feedback relationship between musician and instrument. In presenting the AI within the existing musical context, we can increase the grounding between user and AI as tool and even collaborator. In similar applications of XAI and the arts, and indeed in all applications, this attention to context and using what is already understandable to users will aid the explanation of the underlying systems.

\section{Acknowledgements}
This work was supported by a Queen Mary University of London Research Enabling Fund. The work was undertaken by Banar, Ford, Reed, Zhang, and Armitage under the direction of Bryan-Kinns and with the advice of Colton.
Banar and Ford are research students at the UKRI Centre for Doctoral Training in Artificial Intelligence and Music, supported jointly by UK Research and Innovation (grant number EP/S022694/1) and Queen Mary University of London.
Reed is funded by a Principal Electronic Engineering and Computer Science Studentship from Queen Mary University of London. Yixiao is a research student at the UKRI Centre for Doctoral Training in Artificial Intelligence and Music, supported jointly by the China Scholarship Council and Queen Mary University of London.


\bibliography{references}

\end{document}